\documentclass[10pt,twocolumn,letterpaper]{article}

\usepackage{iccv}
\usepackage{times}
\usepackage{epsfig}
\usepackage{graphicx}
\usepackage{amsmath}
\usepackage{amssymb}
\usepackage{subfigure}
\usepackage{mathrsfs}
\usepackage{booktabs}
\usepackage{multirow}
\usepackage{array}
\usepackage{authblk}

\usepackage{mathrsfs,amsthm,amsmath,amssymb}

\newcommand{\PreserveBackslash}[1]{\let\temp=\\#1\let\\=\temp}
\newcolumntype{C}[1]{>{\PreserveBackslash\centering}p{#1}}
\newcolumntype{R}[1]{>{\PreserveBackslash\raggedleft}p{#1}}
\newcolumntype{L}[1]{>{\PreserveBackslash\raggedright}p{#1}}

\usepackage[pagebackref=true,breaklinks=true,letterpaper=true,colorlinks,bookmarks=false]{hyperref}

\iccvfinalcopy 

\def\iccvPaperID{*} 

\begin{document}

\title{Range Adaptation for 3D Object Detection in LiDAR}

\author[1]{Ze Wang}
\author[2]{Sihao Ding}
\author[2]{Ying Li}
\author[2]{Minming Zhao}
\author[2]{Sohini Roychowdhury}
\author[2]{Andreas Wallin}
\author[1]{Guillermo Sapiro}
\author[1]{Qiang Qiu}

\affil[1]{Duke University \authorcr {\tt\small \{ze.w, guillermo.sapiro, qiang.qiu\}@duke.edu}}

\affil[2]{Volvo Car Technology USA \authorcr {\tt\small \{sihao.ding, ying.li.5, minming.zhao, sohini.roy.chowdhury, andreas.wallin1\}@volvocars.com}}
\maketitle

\begin{abstract}

LiDAR-based 3D object detection plays a crucial role in modern autonomous driving systems. LiDAR data often exhibit severe changes in properties across different observation ranges. 
In this paper, we explore cross-range adaptation for 3D object detection using LiDAR, i.e., far-range observations are adapted to near-range.
This way, far-range detection is optimized for similar performance to near-range one.
We adopt a bird-eyes view (BEV) detection framework to perform the proposed model adaptation. 
Our model adaptation consists of an adversarial global adaptation, and a fine-grained local adaptation.  
The proposed cross-range adaptation framework is validated on three state-of-the-art LiDAR based object detection networks, 
and we consistently observe performance improvement on the far-range objects,
without adding any auxiliary parameters to the model. 
To the best of our knowledge, this paper is the first attempt to study cross-range LiDAR adaptation for object detection in point clouds.
To demonstrate the generality of the proposed adaptation framework, experiments on more challenging cross-device adaptation are further conducted, and a new LiDAR dataset with high-quality annotated point clouds is released to promote future research.

\end{abstract}

\section{Introduction}
\label{intro}
Modern autonomous driving systems rely heavily on accurate and robust perception, where LiDAR-based 3D object detection plays an indispensable role.
Object detection in point clouds has appealing advantages such as accurate distance encoded in points and scale consistency in a 3D space.
Several recent 3D object detection methods report promising results on public LiDAR datasets \cite{yan2018second,zhou2018voxelnet}.
However, within a point cloud, the density of points degrades significantly with the distance to the sensor, leading to inferior performance for far-range objects. Perception of objects in long ranges is crucial for planning ahead while driving, in particular in highways, and new LiDAR technology is being developed to measure far away objects.

In this paper, addressing object detection in point clouds, we propose to perform cross-range adaptation for deep model learning, 
obtaining for far-range objects similar performance to near-range detection
While domain adaptation on optical images has been widely studied, it is still a highly challenging task to perform adaptation on point clouds.
First, the un-ordered and irregular structure of point clouds differs significantly from the gridded structure of optical images, which prevents methods in the image space,  e.g., cGAN based methods, from being directly utilized on point clouds. 
Second, certain unique properties of point clouds, e.g., the scale consistency, bring appealing advantages over optical images, and they need to be fully exploited.

Instead of directly performing adaptation on raw point cloud, we propose model adaptation to be applied on the intermediate layer of a deep network for point cloud object detection.
In the proposed framework, a key step is to align cross-range gridded features at an intermediate layer of a deep network, so that the parameters in the preceding layers are tuned at a low cost to handle range shifts, while the subsequent layers remain shared. 
Specifically, we use near-range areas as the source domain, and improve the feature and the detection accuracy of far-range areas which serve as the target domain. 

We adopt combinations of local and global adaptation.
For the global adaptation,  we adopt adversarial learning to align the feature in the network feature space.
Such methods have delivered outstanding performances on image-based object classification and segmentation. However, as observed for images, adaptation results depend heavily on how complicated the task is. For example,  satisfactory results are obtained for adapting digit images, while limited performance is observed for sophisticated scenarios like image segmentation. These observations motivate us to explore beyond adversarial learning and exploit the properties of point cloud for fine-grained adaptation as detailed next.

Beyond the global adaptation, 
we should note that
while the size of an object varies with distance in optical images, it stays constant in a point cloud.
By exploiting such scale consistency property of point clouds, we propose to mine in the point cloud space for matched local regions across the source and target ranges, and then perform fine-grained local adaptation in the corresponding feature space.

We perform extensive experiments on public 3D object detection datasets and methods. The obtained results validate the proposed framework as an effective method for point cloud range adaptation, including more challenging cross-device adaptation. Beside the superiority on performance, our method does not introduce any auxiliary network layers for the detection model, which enables the adapted object detector to run at the same speed and memory consumption as the original model but with significantly superior detection accuracy.

Our contributions are summarized as follow:
\begin{itemize}
	\item We propose cross-range adaptation to significantly improve LiDAR-based far-range object detection.
	
	\item We combine fine-grained local adaptation and adversarial global adaptation for 3D object detection models.

	\item To the best of our knowledge, this work is the first attempt to study adaptation for 3D object detection in point clouds.
	
	\item We release a new LiDAR dataset with high-quality annotated point clouds to promote future research on object detection and model adaptation for point clouds.
\end{itemize}

\section{Related Work}
\subsection{3D Object Detection}
Object detection in point clouds is an intrinsically three dimensional problem. As such, it is natural to deploy a 3D convolutional network for detection, which is the paradigm of several early works \cite{engelcke2017vote3deep,li20173d}. While providing a straightforward architecture, these methods are slow; e.g. Engelcke \textit{et al.} \cite{engelcke2017vote3deep} require 0.5s for inference on a single point cloud. Most recent methods improve the runtime by projecting the 3D point cloud either onto the ground plane \cite{chen2017multi,ku2018joint} or the image plane \cite{li2016vehicle}. In the most common paradigm the point cloud is organized in voxels and the set of voxels in each vertical column is encoded into a fixed-length, hand-crafted feature to form a pseudo-image which can be processed by a standard image detection architecture. Some notable works here include MV3D \cite{chen2017multi}, AVOD \cite{ku2018joint}, PIXOR \cite{yang2018pixor}, and Complex-YOLO \cite{simon2018complex}, which all use variations on the same fixed encoding paradigm as the first step of their architectures. The first two methods additionally fuse the lidar features with image features to create a multimodal detector. The fusion step used in MV3D and AVOD forces them to use two-stage detection pipelines, while PIXOR and Complex-YOLO use single stage pipelines.
Some recent progresses have significantly improved the both the accuracy and the speed on LiDAR-only object detection. SECOND \cite{yan2018second} adopt sparse convolutional layers that process 3D feature at faster speed with much less memory consumption.
PointPillars \cite{lang2018pointpillars} adopt a novel point cloud encoding layer that enable a high speed and high quality transformation from un-ordered points to gridded representations.

\subsection{Domain Adaptation}
Recently, we have witnessed great progress on domain adaptation for deep neural networks. The achievements on deep domain adaptation generally follow two directions.
The first direction is to do domain adaptation in a single network, where parameters are shared across domains, while the network is forced to produce domain-invariant features by minimizing additional loss functions in the network training \cite{ganin2016domain,long2015learning,long2016deep,long2016unsupervised,tzeng2014deep}. 
The additional losses are imposed to encourage similar features from source and target domains.
The Maximum Mean Discrepancy (MMD) \cite{gretton2007kernel} emerged as a popular metric of domain distance, and was adopted in \cite{tzeng2014deep} as a domain confusion loss to encourage small distance between the source and the target domain final features in a unified network structure and to prevent the network from overfitting to the source domain.
The MK-MMD \cite{gretton2012optimal} is adopted in \cite{long2015learning} as the discrepancy metric between the source and the target domains.
Beyond the first order statistic, second-order statistics are utilized in \cite{hoffman2014lsda}.

In addition to the hand-crafted distribution distance metrics, many recent efforts \cite{ganin2016domain,tzeng2015simultaneous} resort to adversarial training by applying a feature discriminator that is trained alternatively with the main network to do a binary classification of distinguishing the features from the source and the target domain. The main network is trained to fool the feature discriminator so that domain-invariant features contain no information helping the discriminator to decide which domain the features come from.

\section{Cross-range Adaptation}
\begin{figure}[t]
	\centering
	\subfigure[Original image.]{
		\includegraphics[width=1\linewidth]{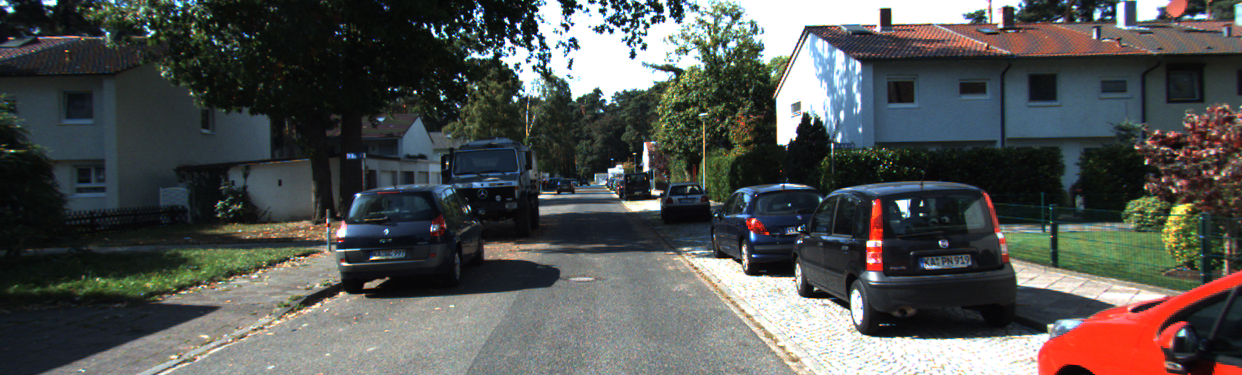}
	}\\
	\subfigure[Image with projected points.]{
		\includegraphics[width=1\linewidth]{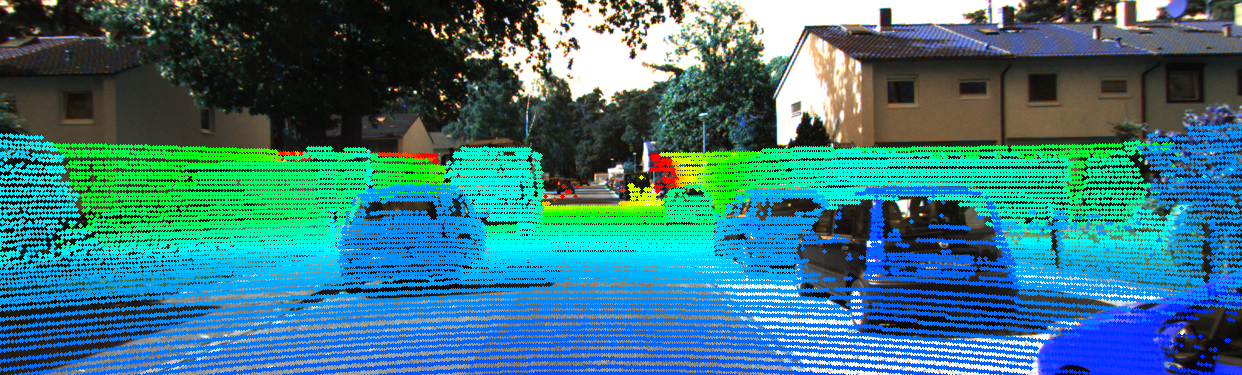}
	}
	\\
	\subfigure[Distance: 8m  Number of points: 1305]{
		\includegraphics[width=0.30\linewidth]{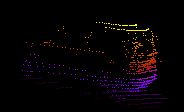}
	}
	\subfigure[Distance: 13m Number of points: 606]{
		\includegraphics[width=0.29\linewidth]{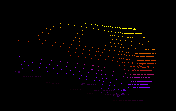}
	}
	\subfigure[Distance: 26m Number of points: 167]{
		\includegraphics[width=0.326\linewidth]{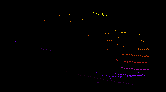}
	}
	\caption{Illustration of the point density. (b) is generated by projecting the point cloud onto the image, which clearly shows that far-range objects are only covered by a very small amount of points. We further select three objects in different distances and plot the points in the corresponding 3D boxes in (c) (d) and (e).}
	\label{fig:point}
\end{figure}

While training a deep network for 3D detection,
we usually have significantly more near-range training samples, which dominates the training loss.
As shown in Figure~\ref{fig:point}, near-range objects are represented with significantly denser points than far-range ones.
Thus, the obtained model usually exhibits superior near-range detection performance, but poor generalization to far-range detection, which needs to be addressed, as self-driving moves to highways and as new sensors with far-range capabilities are being developed.

A point cloud is usually represented by a set of unordered points with continuous values $(x, y, z)$ denoting the cartesian coordinates in 3D space, and optional additional values carrying other physical properties, e.g., reflection value $r$ in the KITTI dataset \cite{geiger2012we}.
Directly processing point clouds using off-the-shelf image-based object detection methods is sub-optimal since point clouds are essentially irregular and unordered, which are not suitable to be processed directly using convolutional neural networks designed for gridded features.
Transforming point clouds to an evenly spaced grid representation is usually adopted for object detection in LiDAR \cite{lang2018pointpillars,yan2018second,zhou2018voxelnet}.

In a gridded feature space, our proposed objective is to align features across different ranges, at a chosen intermediate layer of a 3D object detection network. 
Such chosen layer is referred to as the \emph{aligned layer}. 
Layers preceding such aligned layer, addressed as \emph{adapted layers},  are tuned to
encourage far-range observed objects to produce consistent features as similar objects observed at a near range.
Layers after the aligned layer, address as \emph{shared layers}, determine detection results based on the aligned features.\

\begin{figure}[]
	\centering
	\includegraphics[width=1.0\linewidth]{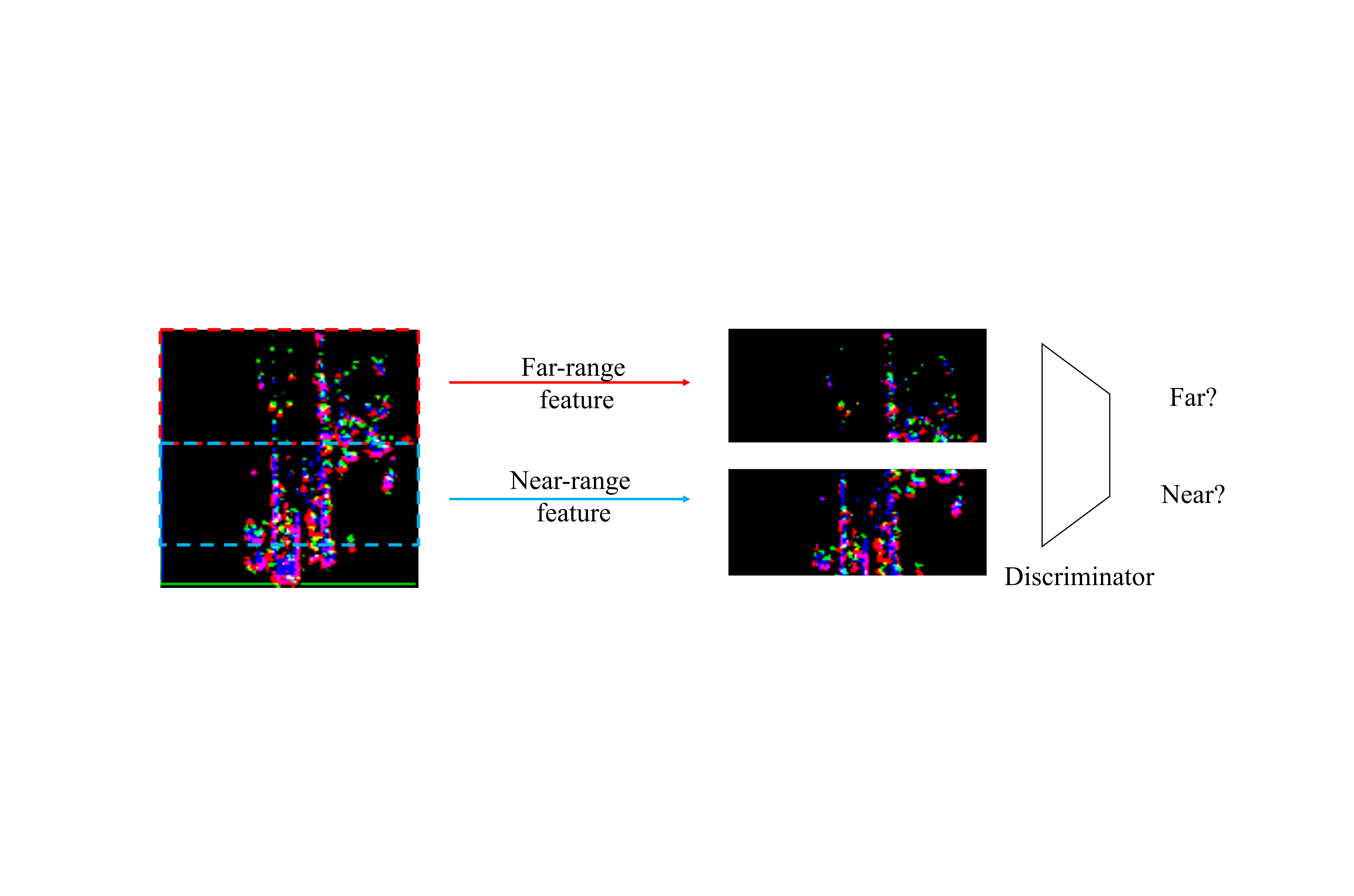}\\
	\caption{Illustration of the global adaptation for cross-range. The source and the target domain are the near-range feature and the far-range feature, respectively. A feature discriminator is applied to promote range-invariant features. Note that we  leave out the area too close to the sensor where unique patterns exist, e.g., the shadow of the data collecting car, that can lead to a quick overfitting to the feature discriminator.
	}
	\label{fig:crossrange_global} 
\end{figure}

To improve the generalization to far-range object detection, we apply both adversarial global adaptation and fine-grained local adaptation.
For global adaptation, as shown in Figure~\ref{fig:crossrange_global}, we set the source domain and the target domain to be near-range features and far-range features of the adapted layer, and use a feature discriminator to promote consistent feature appearance across domains as discussed in Sec.~\ref{aga}.
we then apply an attention mechanism to further align far-range features to near-range features of similar objects, exploiting the unique invariance of point clouds, which is detailed in Section~\ref{fg}. 
The loss of both the global and local adaptations are jointly propagated back to all the layers preceding to the adapted layer.
Note that the proposed framework introduces no additional auxiliary parameters to a deep model.

\subsection{Adversarial Global Adaptation}
\label{aga}

After transforming point clouds to an evenly spaced grid representation, 
adversarial training can be adopted for cross-range adaptation.
Specifically, a feature discriminator is imposed at the aligned layer
to tune parameters at all preceding layers,
so that features from far-range objects become as if they come from near-range ones.
The feature discriminator is implemented as a classifier $C$, which takes both near-range and far-range features as inputs, and is trained to identify which range each feature comes from.
The loss function of the discriminator is expressed as
\begin{equation*}
\begin{aligned}
\mathscr{L}_C(y_n) = - \frac{1}{N} \sum_{N}^{1}[y_n \log(\hat{y}_n) + (1-y_n)\log(1-\hat{y}_n)],
\end{aligned}
\end{equation*}
where $N$ is the total number of features. $y_n \in [0, 1]$ indicates a range label with `0' for far-range and `1' for near-range, and $\hat{y_n}$ is the predication from the feature discriminator.
The parameters of adapted layers and the feature discriminator are updated alternatively. The loss of the detection network becomes from $\mathscr{L}_D$ to $\mathscr{L}_D(1 - y_n)$, which encourages the adapted layers to produce unified features to fool the feature discriminator.
Following \cite{isola2017image}, we adopt a patch-based discriminator, which only penalizes structure at the scale of feature patches.

\subsection{Fine-grained Local Adaptation}
\label{fg}
The above adversarial adaptation provides a global feature alignment, 
a fine-grained local adaptation is proposed to further improve far-range performance. 
In point clouds, an object has a consistent scale, regardless of viewing angles and positions.
In other words, LiDAR observation patterns in the far-range areas
repeat in near-range areas for similar objects but with much denser points. 
Thus, we can mine region pairs of similar patterns to further perform region-based feature adaptation.
Note that similar tasks become significantly harder for images, as we need to simultaneously handle changes in scales, viewing angles, illumination, etc.
This is therefore a unique characteristic of point clouds in general, and LiDAR in particular, we here explicitly exploit.

Given object annotations, during training, we divide objects in each mini-batch into two groups based on the range associated, i.e., a far-range group $ \mathcal{F}$ for objects beyond a range threshold, and a near-range group $\mathcal{N}$ for objects within the range threshold. 
To further encourage a far-range object to share consistent features, at the aligned layer, as similar objects at a near range, we compute the targeted feature of a far-range object as 
a weighted average of all object features in the near-range group. The weight is determined based on the object similarity in the original point cloud space.
Specifically, each object is represented as  $\{\mathbf{o}_i, \mathbf{f}_i\}$, where $\mathbf{o}_i$ denotes its representation in a point cloud space parametrized by the width $w$, height $h$, and yaw-angle $r$;
and $\mathbf{f}_i$ denotes its feature at the aligned layer of a deep network.
Given a far-range object $\{\mathbf{o}_t, \mathbf{f}_t\}$, its targeted feature $\hat{\mathbf{f}_t}$ is determined as,
\begin{equation} \label{tfeature}
\hat{\mathbf{f}_t} = \sum_{i\in \mathcal{N}} w_{it} \mathbf{f}_i, 
\end{equation}
and the weight $w_{it}$ is computed as,
\begin{equation} \label{weight}
w_{it} = \frac{e^{|\mathbf{o}_i - \mathbf{o}_t |}}{\sum_{j\in \mathcal{N}}{e^{|\mathbf{o}_j - \mathbf{o}_t |}}}.
\end{equation}
The final optimization loss then becomes
\begin{equation} \label{loss}
\mathscr{L}_l = \sum_{t \in \mathcal{F}} ||  \mathbf{f}_t - \hat{\mathbf{f}_t} || ^2,
\end{equation}  
which minimizes the distance to the corresponding target feature for each far-range object.
Note that we cut off the gradient propagates through $\hat{\mathbf{f}_t}$ to prevent the network from aligning cross-range features by degrading near-range features.

\begin{figure*}[t]
	\centering
	\includegraphics[width=1.0\linewidth]{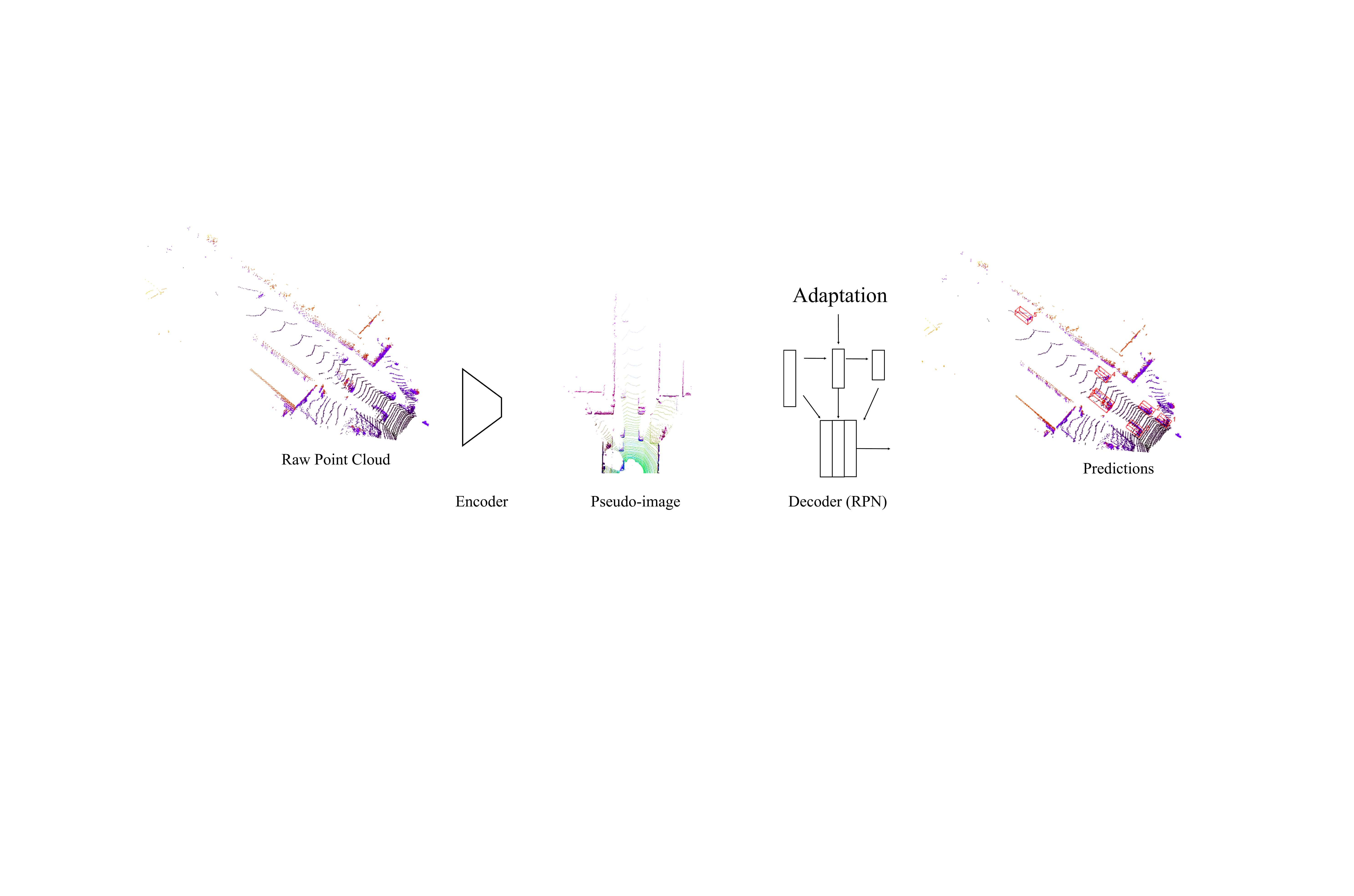}\\
	\caption{Overview of the two-stage BEV object detection pipeline. Different methods apply different techniques to transfer un-ordered point clouds to pseudo-images in either parametric or non-parametric ways. A decoder which is usually formed as a variant of classic detection networks is then applied to generate the final prediction. Our adaptation methods are performed at an intermediate layer of the decoder stage.}
	\label{fig:model} 
\end{figure*}

\section{Experiments}
In this section, we present experiments to validate the effectiveness of the proposed framework.

\subsection{3D Object Detection Methods}
We evaluate the proposed adaptation methods on several state-of-the-art BEV object detection frameworks. Despite significantly different network architectures adopted in these frameworks, each can be briefly described as a two-stage network as shown in Figure~\ref{fig:model}, where the first stage is to transfer the un-ordered and irregular structure into a pseudo-image representation, and a standard object detection head is then applied for 3D box prediction as the second stage. Note that the first stage does not have to be a parametric network, e.g., for Complex-YOLO \cite{simon2018complex}, the pseudo-image is generated using predefined rules without learning.

We select three network architectures as baseline models in the experiments, and 
demonstrate that the detection performance of all three models can be significantly improved with the proposed adaptation method, without any additional parameters to the models.
Complex-YOLO \cite{simon2018complex} is selected as it is an effective method that uses hand-craft rules to generate the pseudo-image. VoxelNet \cite{zhou2018voxelnet} is selected as it is a powerful network architecture that motivated several follow-up methods. SECOND \cite{lang2018pointpillars} is selected as it reports at the moment the best detection performance.
These three network architectures share a similar detection pipeline as shown in Figure~\ref{fig:model}. 
In the encoder stage, Complex-YOLO generates a pseudo-image using predefined rules, SECOND and VoxelNet adopt operations on local regions, e.g., the VFE layers and the sparse convolutional layers. 
Such operations are not suitable for performing adaptation since they are all local operations that operate on a small region on the feature map.
Therefore, we propose to apply the adaptation on the intermediate layer of the decoder stage as shown in Figure~\ref{fig:model}, since the corresponding layer has both strong semantic encoded and compact feature size.

\begin{figure}[]
	\centering
	\includegraphics[width=0.8\linewidth]{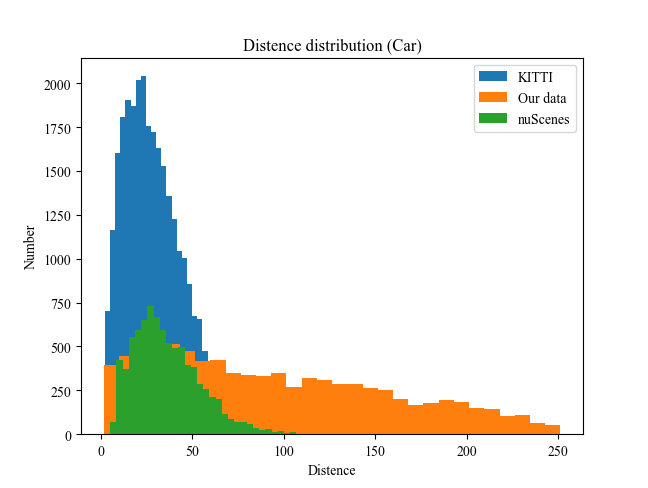}\\
	\caption{Histograms for the distance of cars to the sensor for all the three datasets. We choose the car class for illustration.
	}
	\label{fig:gra} 
	\vspace{-5mm}
\end{figure}

\begin{table*}[t]
	\begin{center}
		\begin{tabular}{c | c|C{1.5cm}|C{1.5cm}|C{1.5cm}|C{1.5cm}|C{1.5cm}|C{1.5cm}}
			\toprule
			\multirow{2}*{} & \multirow{2}*{Methods}& \multicolumn{3}{c|}{Near-range (0-40m)} & \multicolumn{3}{c}{Full-range (0-70m)}\\
			\cline{3-8}
			~&~ & Easy & Moderate & Hard & Easy & Moderate & Hard\\
			
			\midrule
			\multirow{6}*{\rotatebox{90}{3D AP}}
			&Complex-YOLO (w/o) & 85.42& 76.33 & 69.12 &85.24 & 73.51 & 68.33 \\
			~&Complex-YOLO (w)   & 85.90& 77.01 & 72.98  &85.89 & 77.02 & 71.43 \\
			\cmidrule{2-8}
			~&VoxelNet (w/o) & 88.12 & 77.12 & 75.42 & 87.98 & 76.12 & 74.42 \\
			~&VoxelNet (w)  & \textbf{89.04} & 78.12 & 76.01 & \textbf{89.02} & 77.98 & 75.82 \\
			\cmidrule{2-8}
			~&SECOND (w/o) & 88.28& 85.21& 77.57& 88.07& 77.12& 75.27\\
			~&SECOND (w) & 88.81& \textbf{85.84}& \textbf{78.01} & 88.80& \textbf{78.31}& \textbf{76.16}\\
			
			\midrule
			\multirow{6}*{\rotatebox{90}{2D AP}}
			
			&Complex-YOLO (w/o)& 90.62 & 88.89 & 87.12 & 90.48 & 88.48 & 86.76\\
			~&Complex-YOLO (w)  & 90.62 & 88.91 & 87.12 & 90.61 & 88.82 & 88.24\\
			\cmidrule{2-8}
			~&VoxelNet (w/o)  & 90.25 & 89.98 & 89.15 & 89.46& 87.90& 87.72\\
			~&VoxelNet (w)    & 90.28 & 90.02 & 89.65 & 90.07& 89.19& 88.42\\
			\cmidrule{2-8}
			~&SECOND (w/o)& 90.77 & 90.32 & 89.15 & 90.26& 89.19& 88.08\\
			~&SECOND (w) & \textbf{90.78} & \textbf{90.40}& \textbf{89.94} &\textbf{90.78}& \textbf{89.69}& \textbf{88.83}\\
			
			\midrule
			\multirow{6}*{\rotatebox{90}{BEV AP}}
			&Complex-YOLO (w/o) & 89.72 & 88.99 & 87.66 & 89.45& 80.90& 79.49 \\
			~&Complex-YOLO (w)   & 89.92 & 89.05 & 88.08 & 89.64& 82.01& 81.85 \\
			\cmidrule{2-8}
			~&VoxelNet (w/o) & 89.72 & 88.99 & 87.66 & 89.72& 87.37& 79.23 \\
			~&VoxelNet (w)   & 89.74 & 89.12 & \textbf{88.18} & 89.72& \textbf{88.17}& 80.02\\
			\cmidrule{2-8}
			~&SECOND (w/o)& 88.62& 87.81 & 86.39 & 88.62& 86.25& 86.21\\
			~&SECOND (w)& \textbf{90.34}& \textbf{89.37}& 87.67 & \textbf{90.32} & 87.82 & \textbf{87.51}\\
			
			\bottomrule
		\end{tabular}
		\vspace{3mm}
		\caption{Average precisions. Three metrics including 3D bounding box, 3D bounding box, and BEV are reported. Best results are marked in bold.}
		\label{tab:bev}
	\end{center}
\end{table*}

\begin{table*}
	\begin{center} 
		\begin{tabular}{c|c|c|c|c|c|c|c|c|c}
			\toprule
			\multirow{2}*{Methods}& \multicolumn{3}{c|}{3D AP}& \multicolumn{3}{c|}{2D AP} & \multicolumn{3}{c}{BEV AP} \\
			\cline{2-10}
			~& Easy & Mod. & Hard & Easy & Mod. & Hard& Easy & Mod. & Hard \\	
			\midrule 
			SECOND (w/o) & 0.0 & 5.52 & 5.52& 0.0& 9.09 & 12.30 & 0.0 & 10.91 & 11.76\\
			SECOND (w) &   0.0 & \textbf{7.02} & \textbf{7.02}& 0.0& \textbf{13.64} & \textbf{13.64} & 0.0 & \textbf{13.22} & \textbf{13.22}\\
			
			\bottomrule
		\end{tabular}
		\vspace{3mm}
		\caption{Far-range (60-70m) only comparisons. Note the for the easy level, the AP is always 0 since there is no easy object in this area.}
		\label{tab:fr}
	\end{center}
\end{table*}

\subsection{Quantitative Results}
We apply the proposed adaptation framework on the above three network models, and report the quantitative results for fair comparisons.
\paragraph{Dataset.}
The experiments for cross-range adaptation are performed on the KITTI benchmark \cite{geiger2012we}. We follow the standard settings as in \cite{chen2017multi} to split the provided 7,481 samples into a training set of 3,712 samples and an evaluation set of 3,769 samples. The evaluation of a detector for KITTI 3D object detection is performed on three levels of difficulty: easy, moderate, and hard. The difficulty assessment is based on the object heights, occlusion, and truncation.

\begin{figure*}[h]
	\centering
	\subfigure[Comparion at 6,000 iterations.]{
		\includegraphics[width=0.23\linewidth]{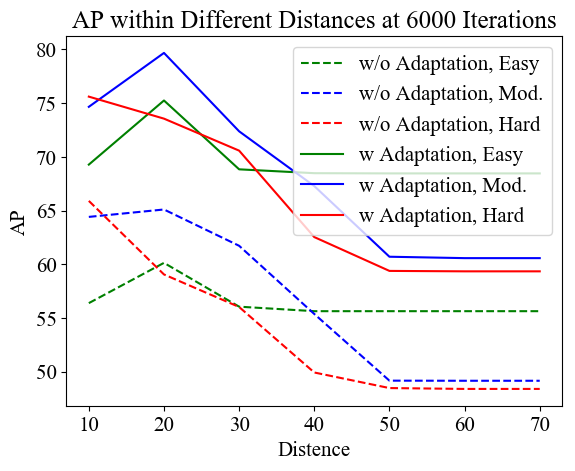}
	}
	\subfigure[Comparion at 12,000 iterations.]{
		\includegraphics[width=0.23\linewidth]{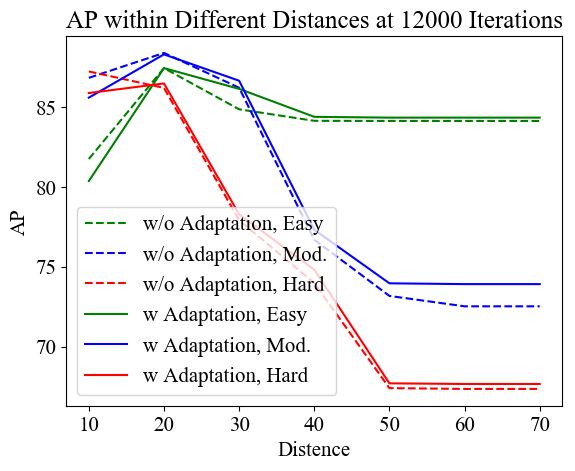}
	}
	\subfigure[Comparion at 21,000 iterations.]{
		\includegraphics[width=0.23\linewidth]{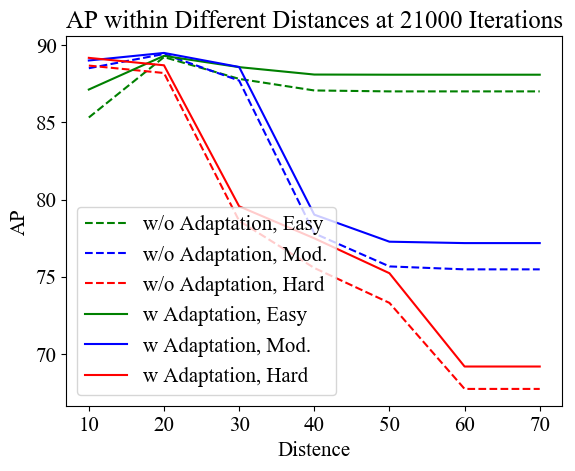}
	}
	\subfigure[Comparion at 30,000 iterations.]{
		\includegraphics[width=0.23\linewidth]{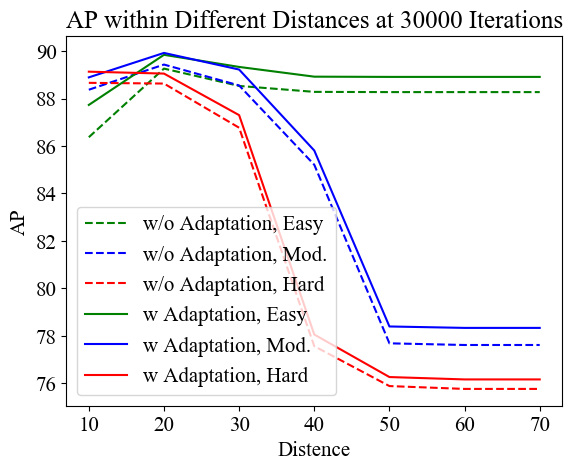}
	}

	\caption{3D bbox AP at different iterations. The cures are obtained by training SECOND with and without the proposed adaptation framework. Training with adaptation gives a significant improvement on AP at the initial stage. It's clearly demonstrated that the proposed adaptation delivers not only a performance improvement on far-range objects, but also a improvement at the near-range objects.}
	\label{fig:d3}
\end{figure*}

\begin{figure*}[]
	\centering
	\subfigure[Reducing false positive.]{
		\includegraphics[width=0.43\linewidth]{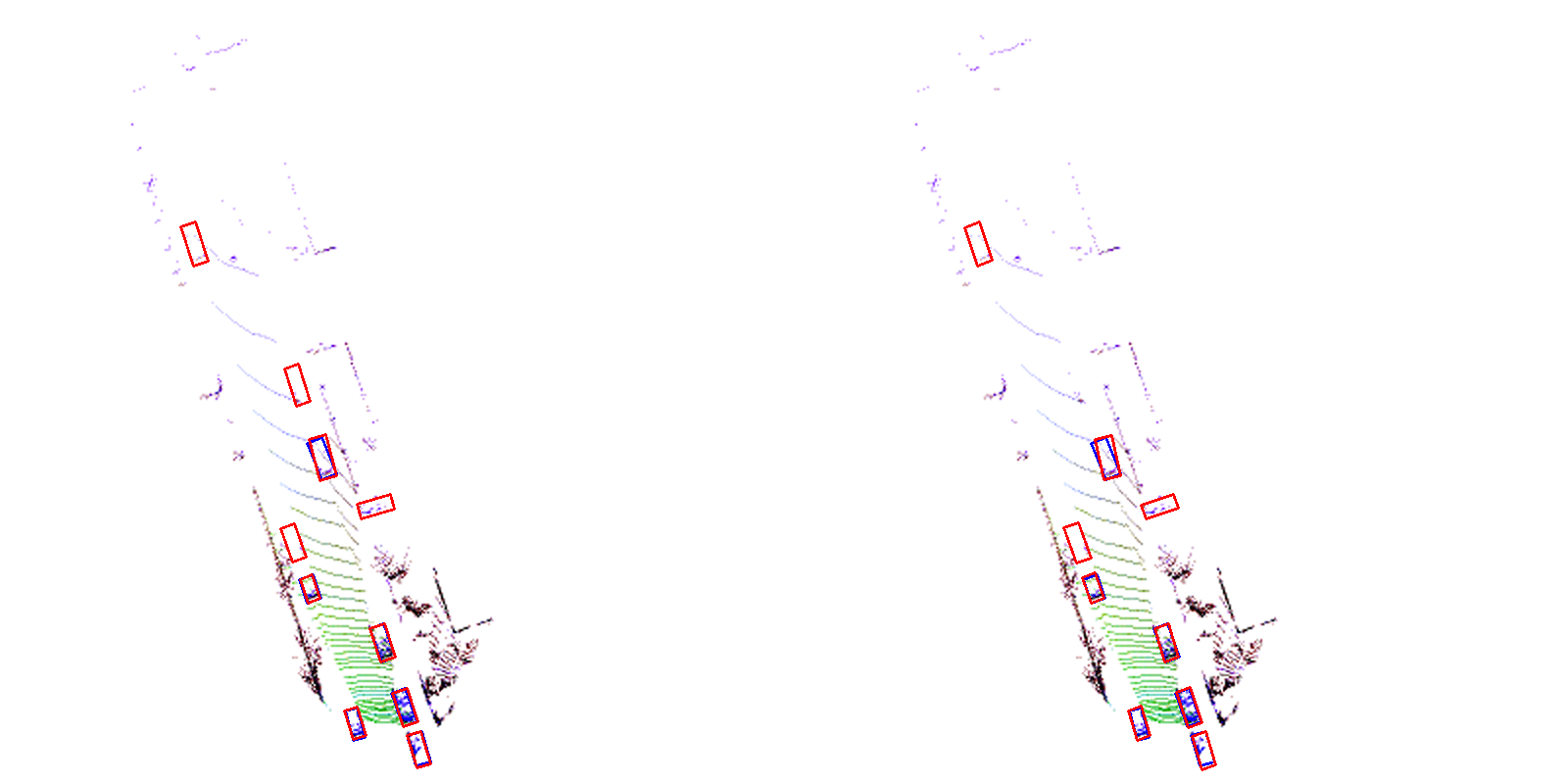}
	}
	\subfigure[Reducing false positive.]{
		\includegraphics[width=0.43\linewidth]{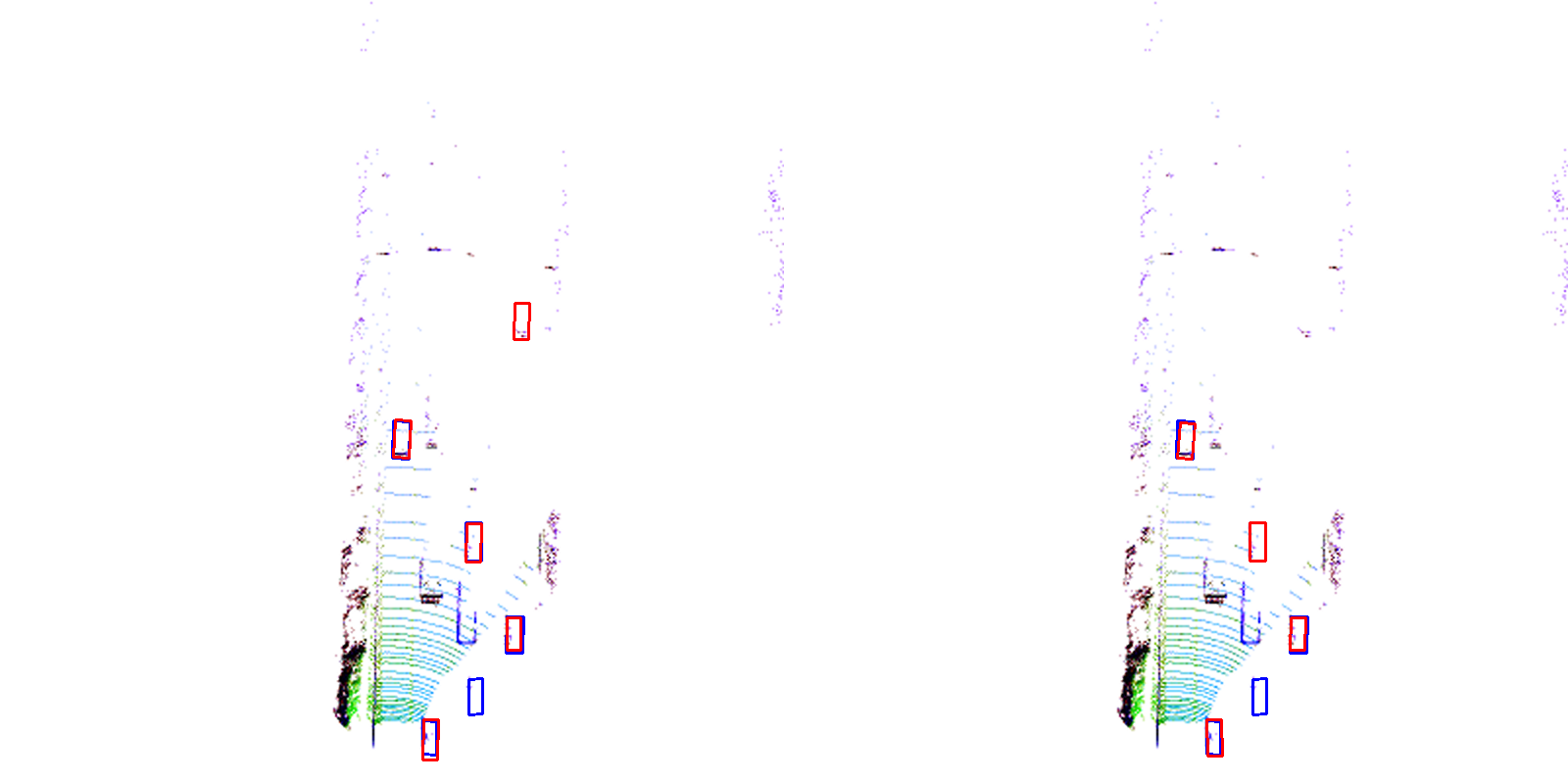}
	}
	\subfigure[Improving far-range recall.]{
		\includegraphics[width=0.43\linewidth]{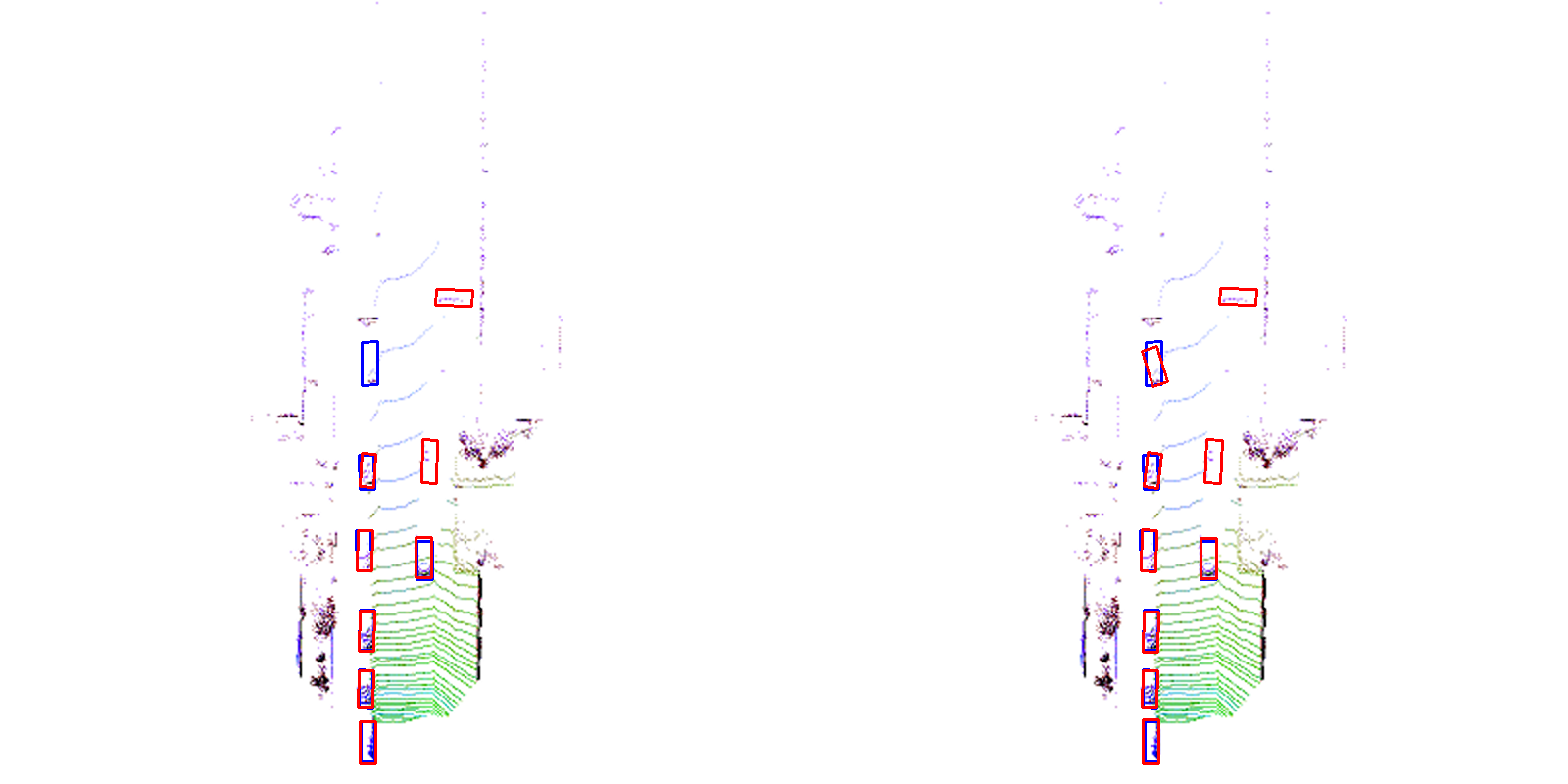}
	}
	\subfigure[Improving far-range recall.]{
		\includegraphics[width=0.43\linewidth]{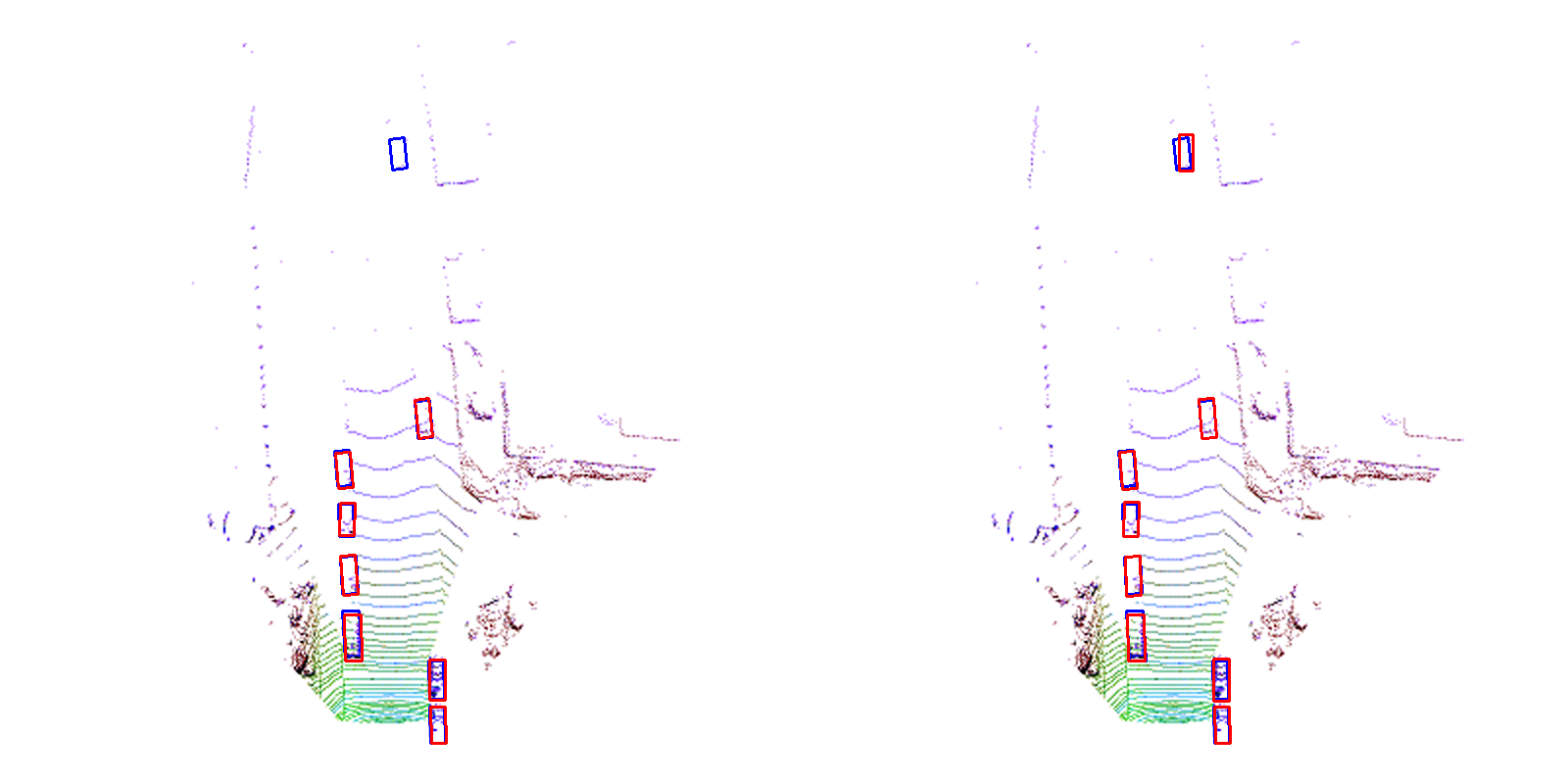}
	}

	\caption{Qualitative results (please zoom in for details). We present paired samples where in each pair, the left image is the result trained on the original network, while the right image is the result with the proposed adaptation framework. We use red and blue boxes to denote detections and ground truth boxes, respectively. It's clearly shown that the proposed framework increases the recall on far-range objects and reduces the false positive in the far-range area.}
	\label{fig:qua}
\end{figure*}

\paragraph{Metrics.}
Following the official KITII evaluation detection metrics, we report average precision (AP) on 2D box, bird's eye view (BEV), and 3D box.
The results on 2D box are computed by projecting the 3D boxes on to the image planes, and calculating the average precision in 2D space.
Note that the 3D box precision is the primary metric in our work.
\paragraph{Implementations.}
Since there is no official implementation for Complex-YOLO \cite{simon2018complex} and VoxelNet \cite{zhou2018voxelnet}, we follow the descriptions in the papers and reimplement the networks. 
For the implementation of SECOND, we directly use the authors' publicly available implementation for a fair comparison.\footnote{https://github.com/traveller59/second.pytorch}
All the experiments are implemented in PyTorch \cite{paszke2017automatic}, and are trained on a single GTX 1080Ti graphic card.

Since all these three networks we select use different configurations for object categories, we only report detection results on cars, as cars are the object with a significantly dominant amount in the dataset for a reliable comparison.
Following \cite{zhou2018voxelnet}, we set the full scale detection range to be $[-3, 1] \times [-40, 40] \times [0, 70.4]$ meters along the Z, Y, X axis, respectively. Note that as clearly indicated in \cite{simon2018complex}, Complex-YOLO neglects the objects that are farther than 40m for the sake of efficiency. We use the same range of $[0, 70.4]$ meters along the X axis across all the experiments for a fair comparison.
Based on the distribution presented in Figure~\ref{fig:gra}, we set the range threshold to be 40m.
We report detection accuracies on near-range 0-40m and full-range 0-90m with and without adaptation, to demonstrate that the proposed framework can improve the far-range detection accuracy without any compromise on the near-range performance. 
The qualitative results performed on the three networks with and without adaptation are presented in Table~\ref{tab:bev}. 
We observe from the results that the proposed adaptation framework not only boosts the far-range performance as shown in Table~\ref{tab:fr}, but also increase the near-range performance at the same time, thus we achieve  a superior full-range detection accuracy in Table~\ref{tab:bev}.

We further plot the AP-distance curves in different training stages in Figure~\ref{fig:d3}. Our framework significantly accelerates the training speed in the early stage as shown in Figure~\ref{fig:d3}(a), and helps the network converge with a better performance (30000 iterations) as shown in Figure~\ref{fig:d3}(d). In the entire training procedure, the proposed framework consistently improves the performance in both the near-range and the far-range areas. We further present a comparison performed on SECOND to validate the performance growth in the far-range only area, the results are presented in Table~\ref{tab:fr}.

\subsection{Qualitative Results}
Qualitative results are presented in Figure~\ref{fig:qua}.
We present paired samples that are generated by SECOND \cite{yan2018second} with and without the proposed adaptation framework, respectively. 
We consistently observe that the proposed framework improves the quality in the far-range area, e.g., the false positives are reduced, and hard positive objects are detected. Meanwhile, the performance in the near-range area remain robust without any degradation.
Note that the networks for presenting the paired examples are trained using exact the same initialization and mini-batches by fixing the random seed for a fair comparison.

\begin{table*}[h]
	\begin{center} 
		\begin{tabular}{c|C{1.5cm}|C{1.5cm}|C{1.5cm}|C{1.5cm}|C{1.5cm}|C{1.5cm}}
			\toprule
			\multirow{2}*{Methods}	 & \multicolumn{3}{c|}{Near-range (0-40m)} & \multicolumn{3}{c}{Full-range (0-70m)}\\
			\cline{2-7}
			~ & Easy & Moderate & Hard & Easy & Moderate & Hard\\
			\midrule
			
			SECOND (w/o) & 88.28& 85.21& 77.57& 88.07& 77.12& 75.27\\
			SECOND (w L) & 88.29& 85.41& 77.58 & 88.23& 77.88& 75.74\\
			SECOND (w G) & 88.62& 85.51& 77.99 & 88.60& 78.02& 75.57\\
			SECOND (w L+G) & \textbf{88.81}& \textbf{85.84}& \textbf{78.01} & \textbf{88.80}& \textbf{78.31}& \textbf{76.16}\\
			
			\bottomrule
		\end{tabular}
		\vspace{3mm}
		\caption{Ablation study. 3D bounding box average precision. L and G denote local adaptation and global adaptation respectively. The results are obtained by running five rounds of comparisons, where an identical random seed is used in each round for each experiment. The presented result is the average of the five round results. The proposed adaptation further improves the best detection model without introducing additional parameters or computation.}
		\label{tab:as}
	\end{center}
\end{table*}

\subsection{Ablation Studies}
We further provide more experiment details and self-comparisons.
All the experiments presented in the ablation studies are conducted on SECOND \cite{yan2018second}. We select SECOND since it reports the best results among the three networks we adopt, and the experiments are implemented based on the publicly available source code online.
Specifically, we validate the contributions of the proposed global and local adaptation by imposing each adaptation method individually on the training of SECOND, and compare the final performances. The results are presented in Table~\ref{tab:as}.

\begin{figure}[]
	\centering
	\vspace{-5mm}
	\subfigure[Sample for KITTI.]{
		\includegraphics[width=0.46\linewidth]{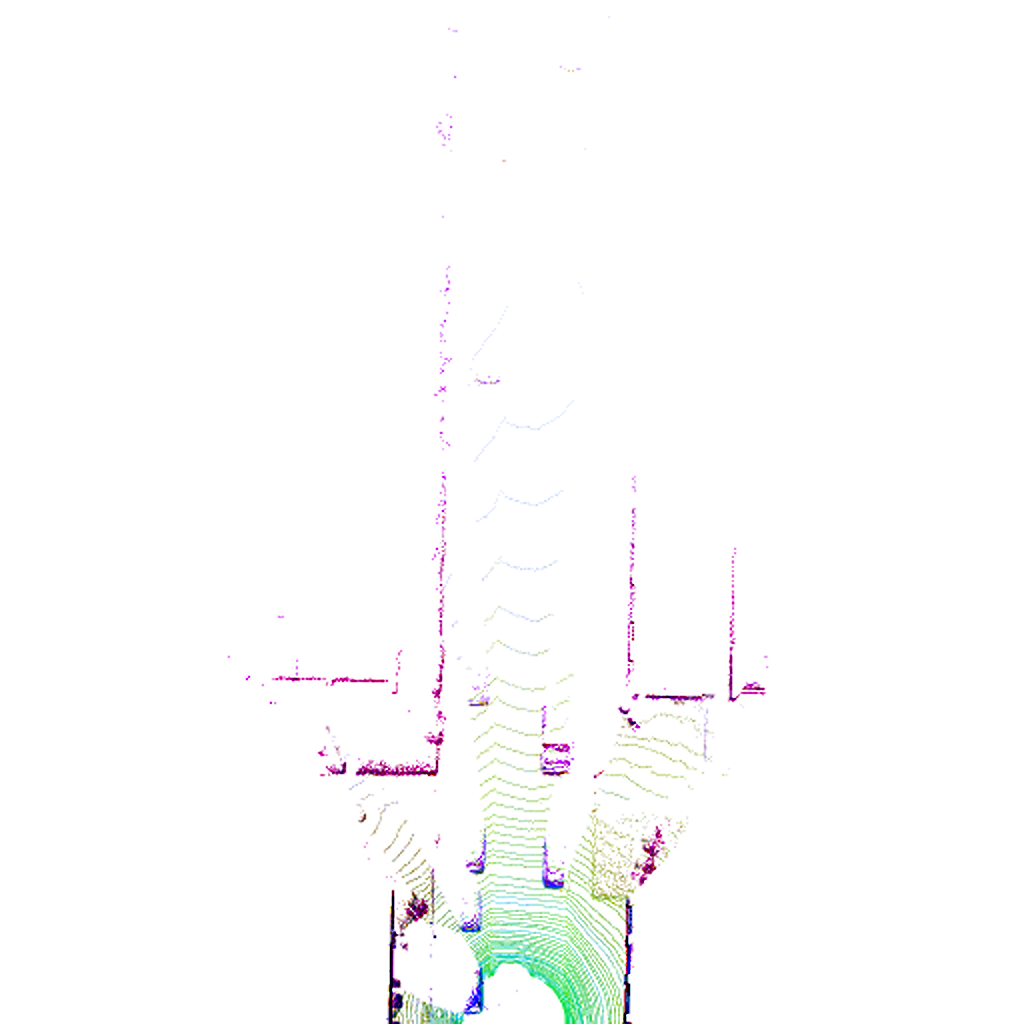}
	}
	\subfigure[Sample for nuScenes.]{
		\includegraphics[width=0.46\linewidth]{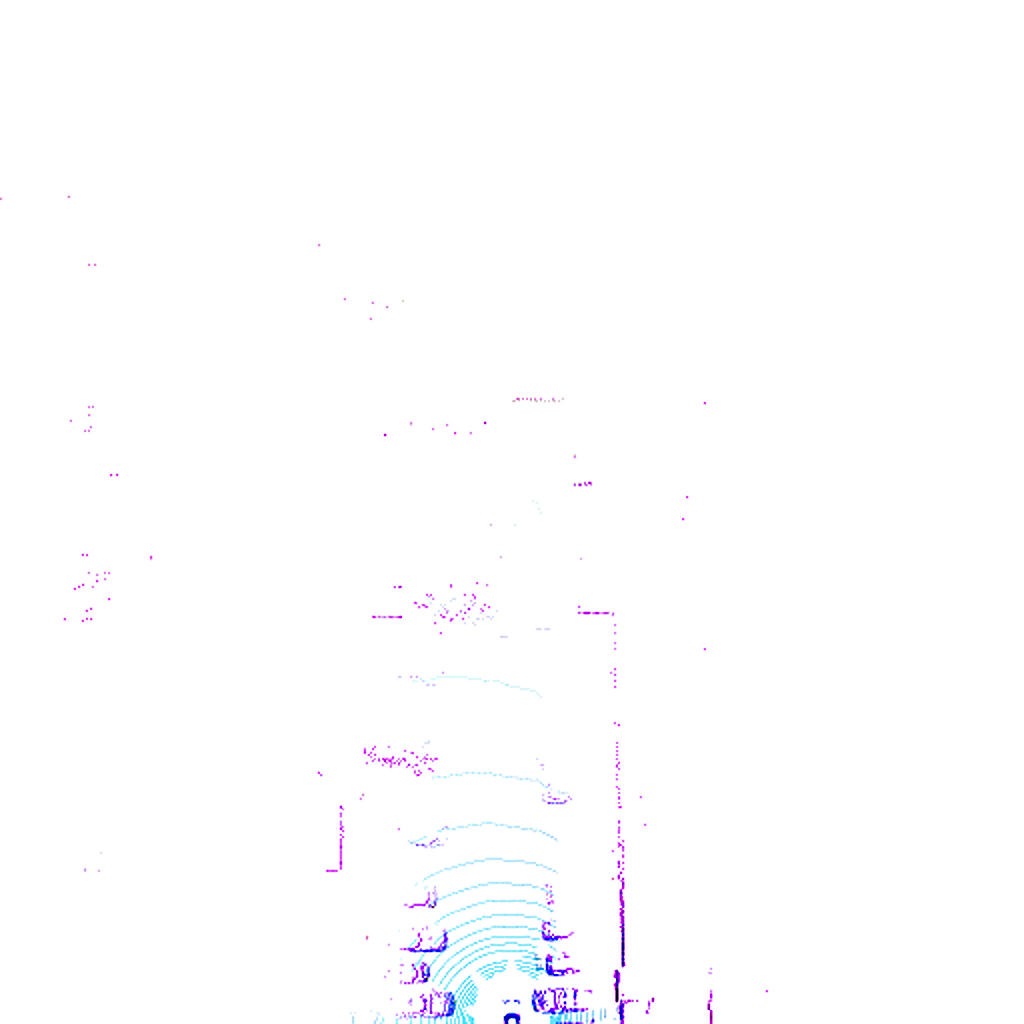}
	}
	\subfigure[Sample for our dataset.]{
		\includegraphics[width=0.46\linewidth]{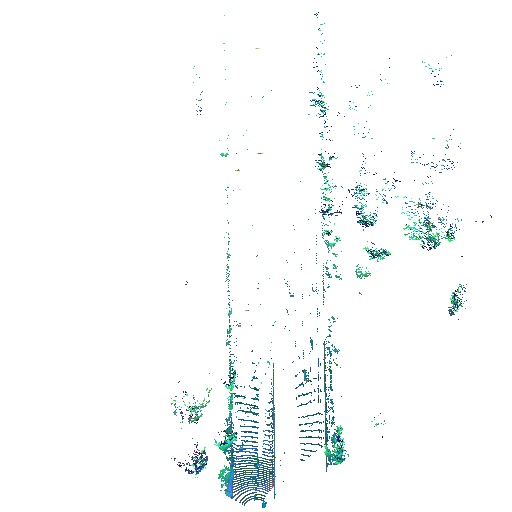}
	}
	\caption{A side-by-side comparison on the point density of samples from KITTI, nuScenes, and our dataset (please zoom in for details). We clearly observe that the nuScenes data has the lowest density, and our data has densest points and longest effective range.}
	\label{fig:ss}
\end{figure}

\begin{table}
	\begin{center} 
		\begin{tabular}{c|cc}
			\toprule
			Dataset  & $\#$ of channels & Effective range \\
			\midrule 
			nuScenes &  32 & 70m \\
			KITTI  & 64 & 120m \\
			Our dataset & 64 & 250m \\
			\bottomrule
		\end{tabular}
		\vspace{3mm}
		\caption{Sensors that used for the collection of different dataset. High density and far effective range make our dataset a strong benchmark for evaluating cross-device adaptation.}
		\label{tab:datasets}
	\end{center}
\vspace{-7mm}
\end{table}

\begin{table}
	\begin{center} 
		\begin{tabular}{c|c}
			\toprule
			Method & AP \\	
			\midrule 
			nuScenes only & 36.2\\
			Joint &  44.3\\
			Joint + Adaptation (w L+G) & \textbf{47.4} \\
			\midrule 
			Our data only & 31.7\\
			Joint &  34.9 \\
			Joint + Adaptation (w L+G) & \textbf{37.7 }\\
			\bottomrule
		\end{tabular}
		\vspace{3mm}
		\caption{Cross-device adaptation. Experiments are performed on two datasets. Joint training with proposed adaptation methods consistently improve the performance on target datasets.}
		\label{tab:cd}
	\end{center}
\vspace{-5mm}
\end{table}

\subsection{Cross-device Adaptation}
The proposed framework on cross-range feature adaptation makes the first step on studying the adaptation in point clouds. Here we take one further step by showing the framework can be extended to, for example, a more challenging cross-device scenario.
The experiment is conducted by introducing other datasets, i.e., nuScenes \cite{nuscenes} 3D object detection dataset and our own dataset,\footnote{The dataset will be publicly available.} which are collected using different devices. As we see from Figure~\ref{fig:gra}, our dataset has a significantly larger range, and adapting to it without heavily re-labeling is critical for its full exploitation.
 
Sensor parameters for each dataset are presented in Table~\ref{tab:datasets}. Our dataset provides the densest point clouds with 128 channels, and farther range with an effective range of 250m.
The significant gap among the parameters of the above sensors makes them a set of perfect comparisons for performing cross-sensor experiments. The significant cross-device gap can be observed in Figure~\ref{fig:ss}.
Although nuScenes provides annotations on all directions, here we only considerate objects that can be seen in the front camera for a fair comparison with KITTI. There are totally 28,742 cars in the 7,481 frames in KITTI training set, and totally 8,426 cars in the 3,977 frames in nuScenes V0.1. In our dataset, there are totally 8,550 cars annotated in 898 frames. For both KITTI and nuScenes datasets, it is clearly shown in Figure~\ref{fig:gra} that the objects in about 50m have a dominant number and only a small number of objects are more than 70m away from the sensor. Our dataset has a clearly more average distribution, and a considerably large amount of objects in the far range (up to 250m).

We use only the initial released proportion of nuScenes V0.1 dataset which contains 3977 frames with annotations.
We split the 3977 frame into a training set with the first 2000 frames, and a testing set with the rest 1977 frames.
For our proposed new dataset, there are totally 898 frames in the initial release, and we use the first 600 frames for training, and the rest for testing.
We use KITTI as the source domain, and conduct two experiments using nuScenes and our dataset as the target domain, respective.
The proposed global and local adaptations are imposed to promote objects in the target dataset to have consistent feature with objects in KITTI.
The experiment is also conducted on the car (`vehicle.car' for nuScenes) class only.
We compare our results with two baselines, training using target dataset only, and training both source and target datasets jointly without adaptation.
We report 3D box AP across the entire range only with no subsets of different difficulties.
The results are presented in Table~\ref{tab:cd}.
Joint training the samples from two datasets with the proposed adaptations delivers the best performance. We hypothesize that the small amount of samples in nuScenes is not enough for training a robust detector. 
Massive data in KITTI dataset and the promoted consistent features improve the robustness for the training on nuScenes and our dataset, so that higher performances are observed on the test sets.

\section{Conclusion and Future Work}
We proposed for the first time a model adaptation for object detection in 3D point clouds.
Specifically, we addressed cross-range and cross-device adaptation using adversarial global adaptation and fine-grained local adaptation. We evaluated our adaptation method on various BEV-based object detection
methods, and demonstrated that the combinations of the global
and local adaptation can significantly improve model detection accuracy without adding any auxiliary
parameters to the model.
Beyond the range and device adaptations here studied,
we will further investigate adaptations under other settings, e.g., adaptation across point clouds collected with different scanning patterns (i.e., Gaussian pattern and uniform pattern),
and developing further adaptation methods that deliver better adaptation results with fewer or even no annotations on target domain.

\section*{Acknowledgments}
Work partially supported by NSF, ARO, ONR, NGA, and gifts from AWS.


{\small
\bibliographystyle{ieee_fullname}
\bibliography{egbib}

\begin{thebibliography}{10}\itemsep=-1pt

\bibitem{nuscenes}
nu{S}cenes dataset.
\newblock \url{https://www.nuscenes.org/}.

\bibitem{chen2017multi}
Xiaozhi Chen, Huimin Ma, Ji Wan, Bo Li, and Tian Xia.
\newblock Multi-view 3d object detection network for autonomous driving.
\newblock In {\em Proceedings of the IEEE Conference on Computer Vision and
  Pattern Recognition}, pages 1907--1915, 2017.

\bibitem{engelcke2017vote3deep}
Martin Engelcke, Dushyant Rao, Dominic~Zeng Wang, Chi~Hay Tong, and Ingmar
  Posner.
\newblock Vote3deep: Fast object detection in 3d point clouds using efficient
  convolutional neural networks.
\newblock In {\em 2017 IEEE International Conference on Robotics and Automation
  (ICRA)}, pages 1355--1361. IEEE, 2017.

\bibitem{ganin2016domain}
Yaroslav Ganin, Evgeniya Ustinova, Hana Ajakan, Pascal Germain, Hugo
  Larochelle, Fran{\c{c}}ois Laviolette, Mario Marchand, and Victor Lempitsky.
\newblock Domain-adversarial training of neural networks.
\newblock {\em The Journal of Machine Learning Research}, 17(1):2096--2030,
  2016.

\bibitem{geiger2012we}
Andreas Geiger, Philip Lenz, and Raquel Urtasun.
\newblock Are we ready for autonomous driving? the kitti vision benchmark
  suite.
\newblock In {\em 2012 IEEE Conference on Computer Vision and Pattern
  Recognition}, pages 3354--3361. IEEE, 2012.

\bibitem{gretton2007kernel}
Arthur Gretton, Karsten~M Borgwardt, Malte Rasch, Bernhard Sch{\"o}lkopf, and
  Alex~J Smola.
\newblock A kernel method for the two-sample-problem.
\newblock In {\em Advances in Neural Information Processing Systems}, pages
  513--520, 2007.

\bibitem{gretton2012optimal}
Arthur Gretton, Dino Sejdinovic, Heiko Strathmann, Sivaraman Balakrishnan,
  Massimiliano Pontil, Kenji Fukumizu, and Bharath~K Sriperumbudur.
\newblock Optimal kernel choice for large-scale two-sample tests.
\newblock In {\em Advances in Neural Information Processing Systems}, pages
  1205--1213, 2012.

\bibitem{hoffman2014lsda}
Judy Hoffman, Sergio Guadarrama, Eric~S Tzeng, Ronghang Hu, Jeff Donahue, Ross
  Girshick, Trevor Darrell, and Kate Saenko.
\newblock Lsda: Large scale detection through adaptation.
\newblock In {\em Advances in Neural Information Processing Systems}, pages
  3536--3544, 2014.

\bibitem{isola2017image}
Phillip Isola, Jun-Yan Zhu, Tinghui Zhou, and Alexei~A Efros.
\newblock Image-to-image translation with conditional adversarial networks.
\newblock In {\em Proceedings of the IEEE Conference on Computer Vision and
  Pattern Recognition}, pages 1125--1134, 2017.

\bibitem{ku2018joint}
Jason Ku, Melissa Mozifian, Jungwook Lee, Ali Harakeh, and Steven~L Waslander.
\newblock Joint 3d proposal generation and object detection from view
  aggregation.
\newblock In {\em 2018 IEEE/RSJ International Conference on Intelligent Robots
  and Systems (IROS)}, pages 1--8. IEEE, 2018.

\bibitem{lang2018pointpillars}
Alex~H Lang, Sourabh Vora, Holger Caesar, Lubing Zhou, Jiong Yang, and Oscar
  Beijbom.
\newblock Pointpillars: Fast encoders for object detection from point clouds.
\newblock {\em arXiv preprint arXiv:1812.05784}, 2018.

\bibitem{li20173d}
Bo Li.
\newblock 3d fully convolutional network for vehicle detection in point cloud.
\newblock In {\em 2017 IEEE/RSJ International Conference on Intelligent Robots
  and Systems (IROS)}, pages 1513--1518. IEEE, 2017.

\bibitem{li2016vehicle}
Bo Li, Tianlei Zhang, and Tian Xia.
\newblock Vehicle detection from 3d lidar using fully convolutional network.
\newblock {\em arXiv preprint arXiv:1608.07916}, 2016.

\bibitem{long2015learning}
Mingsheng Long, Yue Cao, Jianmin Wang, and Michael~I Jordan.
\newblock Learning transferable features with deep adaptation networks.
\newblock {\em arXiv preprint arXiv:1502.02791}, 2015.

\bibitem{long2016deep}
Mingsheng Long, Han Zhu, Jianmin Wang, and Michael~I Jordan.
\newblock Deep transfer learning with joint adaptation networks.
\newblock {\em arXiv preprint arXiv:1605.06636}, 2016.

\bibitem{long2016unsupervised}
Mingsheng Long, Han Zhu, Jianmin Wang, and Michael~I Jordan.
\newblock Unsupervised domain adaptation with residual transfer networks.
\newblock In {\em Advances in Neural Information Processing Systems}, pages
  136--144, 2016.

\bibitem{paszke2017automatic}
Adam Paszke, Sam Gross, Soumith Chintala, Gregory Chanan, Edward Yang, Zachary
  DeVito, Zeming Lin, Alban Desmaison, Luca Antiga, and Adam Lerer.
\newblock Automatic differentiation in pytorch.
\newblock 2017.

\bibitem{simon2018complex}
Martin Simon, Stefan Milz, Karl Amende, and Horst-Michael Gross.
\newblock Complex-yolo: An euler-region-proposal for real-time 3d object
  detection on point clouds.
\newblock In {\em European Conference on Computer Vision}, pages 197--209.
  Springer, 2018.

\bibitem{tzeng2015simultaneous}
Eric Tzeng, Judy Hoffman, Trevor Darrell, and Kate Saenko.
\newblock Simultaneous deep transfer across domains and tasks.
\newblock In {\em Proceedings of the IEEE International Conference on Computer
  Vision}, pages 4068--4076, 2015.

\bibitem{tzeng2014deep}
Eric Tzeng, Judy Hoffman, Ning Zhang, Kate Saenko, and Trevor Darrell.
\newblock Deep domain confusion: Maximizing for domain invariance.
\newblock {\em arXiv preprint arXiv:1412.3474}, 2014.

\bibitem{yan2018second}
Yan Yan, Yuxing Mao, and Bo Li.
\newblock Second: Sparsely embedded convolutional detection.
\newblock {\em Sensors}, 18(10):3337, 2018.

\bibitem{yang2018pixor}
Bin Yang, Wenjie Luo, and Raquel Urtasun.
\newblock Pixor: Real-time 3d object detection from point clouds.
\newblock In {\em Proceedings of the IEEE Conference on Computer Vision and
  Pattern Recognition}, pages 7652--7660, 2018.

\bibitem{zhou2018voxelnet}
Yin Zhou and Oncel Tuzel.
\newblock Voxelnet: End-to-end learning for point cloud based 3d object
  detection.
\newblock In {\em Proceedings of the IEEE Conference on Computer Vision and
  Pattern Recognition}, pages 4490--4499, 2018.

\end{thebibliography}
}

\end{document}